\documentclass[final]{cvpr}

\usepackage{times}
\usepackage{epsfig}
\usepackage{graphicx}
\usepackage{amsmath}
\usepackage{siunitx}
\usepackage{amssymb}
\makeatletter
\@namedef{ver@everyshi.sty}{}
\makeatother
\usepackage{pgfplots}
\usepackage{times}
\usepackage{epsfig}
\usepackage{xspace}
\usepackage{color}
\usepackage{multirow}
\usepackage{graphics}
\usepackage{adjustbox}
\usepackage{subfigure}
\usepackage{amsmath}
\usepackage{algorithmicx,algorithm}
\usepackage{paralist} 
\usepackage{enumitem}
\usepackage{graphicx} 
\usepackage{epstopdf}
\usepackage{booktabs} 

\usepackage[pagebackref=true,breaklinks=true,letterpaper=true,colorlinks,citecolor=blue,linkcolor=blue,bookmarks=false]{hyperref}

\long\def\ignorethis#1{}

\definecolor{Gray}{rgb}{0.35,0.35,0.35}
\definecolor{Blue}{rgb}{0,0.2,0.8}
\definecolor{Red}{rgb}{0.8,0.2,0}
\definecolor{Green}{rgb}{0.0,0.5,0.1}
\definecolor{Gray}{rgb}{0.4,0.4,0.4}

\newlength\paramargin
\newlength\figmargin
\newlength\secmargin

\usepackage{array}
\newcolumntype{L}[1]{>{\raggedright\let\newline\\\arraybackslash\hspace{0pt}}m{#1}}
\newcolumntype{C}[1]{>{\centering\let\newline\\\arraybackslash\hspace{0pt}}m{#1}}
\newcolumntype{R}[1]{>{\raggedleft\let\newline\\\arraybackslash\hspace{0pt}}m{#1}}

\def\eg{e.g.,~}

\def\vs{vs.~}

\graphicspath{{figure/eps/}}

\graphicspath{ {figures/} }

\definecolor{chromeyellow}{rgb}{1.0, 0.65, 0.0}
\definecolor{darkcerulean}{rgb}{0.03, 0.27, 0.49}
\definecolor{darkorange}{rgb}{1.0, 0.55, 0.0}
\definecolor{darkmidnightblue}{rgb}{0.0, 0.2, 0.4}
\definecolor{internationalorange}{rgb}{1.0, 0.31, 0.0}
\definecolor{internationalkleinblue}{rgb}{0.0, 0.18, 0.65}
\definecolor{lightsalmon}{rgb}{1.0, 0.63, 0.48}
\definecolor{mangotango}{rgb}{1.0, 0.51, 0.26}
\definecolor{mayablue}{rgb}{0.45, 0.76, 0.98}
\definecolor{majorelleblue}{rgb}{0.38, 0.31, 0.86}
\definecolor{mediumelectricblue}{rgb}{0.01, 0.31, 0.59}

\newcolumntype{x}[1]{>{\centering\arraybackslash}p{#1pt}}

\newlength\savewidth\newcommand\shline{\noalign{\global\savewidth\arrayrulewidth
		\global\arrayrulewidth 1pt}\hline\noalign{\global\arrayrulewidth\savewidth}}
		
\makeatletter\renewcommand\paragraph{\@startsection{paragraph}{4}{\z@}
	{.5em \@plus1ex \@minus.2ex}{-.5em}{\normalfont\normalsize\bfseries}}\makeatother
	
\usepackage[pagebackref=true,breaklinks=true,colorlinks,bookmarks=false]{hyperref}

\def\cvprPaperID{2476} %
\def\confYear{CVPR 2021}

\begin{document}

\title{Lite-HRNet: A Lightweight High-Resolution Network}

\author{
    Changqian Yu$^{1,2}$
    ~
    Bin Xiao$^2$
    ~
    Changxin Gao$^1$
    ~
    Lu Yuan$^2$
    ~
    Lei Zhang$^2$
    ~ 
    {Nong Sang$^1$}\thanks{Corresponding author.
    This work is done when C. Yu was an intern at Microsoft Research, Beijing, P.R. China.}
    ~
    {Jingdong Wang$^{2*}$}\\
    \normalsize 
    $^1$Key Laboratory of Image Processing and Intelligent Control\\
    \normalsize
    School of Artificial Intelligence and Automation, Huazhong University of Science and Technology\\
    \normalsize
    $^2$Microsoft\\
{\small \tt \{changqian\_yu,cgao,nsang\}@hust.edu.cn, \{bixi,luyuan,leizhang,jingdw\}@microsoft.com
} \\
}

\maketitle
\pagestyle{empty}
\thispagestyle{empty}

\begin{abstract}
We present an efficient high-resolution network, Lite-HRNet, for human pose estimation.
We start by simply applying the efficient shuffle block in ShuffleNet
to HRNet (high-resolution network), yielding stronger performance
over popular lightweight networks, 
such as MobileNet, ShuffleNet, and
Small HRNet.

We find that 
the heavily-used pointwise ($1\times 1$) convolutions 
in shuffle blocks become the computational bottleneck.
We introduce a lightweight unit,
conditional channel weighting,
to
replace costly pointwise ($1\times 1$) convolutions in shuffle blocks.
The complexity of channel weighting is 
linear w.r.t the number of channels
and lower than
the quadratic time complexity for pointwise convolutions.
Our solution learns the weights 
from
all the channels 
and over multiple resolutions
that are readily available in the parallel branches
in HRNet.
It uses the weights as the bridge to 
exchange information 
across channels and resolutions,
compensating the role played by the pointwise ($1\times 1$) convolution.
Lite-HRNet demonstrates superior results on human pose estimation over popular lightweight networks.
Moreover, Lite-HRNet can be easily applied to semantic segmentation task in the same lightweight manner.
The code and models have been publicly
available at \url{https://github.com/HRNet/Lite-HRNet}.

\end{abstract}

\section{Introduction}
\label{sec:intro}
Human pose estimation requires high-resolution representation~\cite{Chen-ICLR-Deeplabv2-2016, Badrinarayanan-PAMI-SegNet-2017, Lin-CVPR-FPN-2017, Wang-TPAMI-HRNet-2019,wu20203d}
to achieve high performance.
Motivated by the increasing demand for model efficiency,
this paper studies the problem of developing efficient high-resolution models 
under computation-limited resources.

Existing efficient networks~\cite{Chen-CVPR-DynamicConv-2020,Chen-ECCV-DyReLU-2020,Yu-ECCV-BiSeNet-2018} 
are mainly designed from two perspectives.
One is to borrow the design from classification networks,
such as MobileNet~\cite{Howard-Arxiv-MobileNet-2017, Howard-ICCV-Mobilenetv3-2019} and ShuffleNet~\cite{Ma-ECCV-Shufflenetv2-2018, Zhang-CVPR-Shufflenet-2018},
to reduce the redundancy in matrix-vector multiplication,
where convolution operations dominate the cost.
The other is to mediate the spatial information loss 
with various tricks, such as encoder-decoder architectures~\cite{Badrinarayanan-PAMI-SegNet-2017, Lin-CVPR-FPN-2017}, 
and multi-branch architectures~\cite{Yu-ECCV-BiSeNet-2018, Zhao-ECCV-ICNet-2018}.

We first study a naive lightweight network 
by simply combining the shuffle block in ShuffleNet
and the high-resolution design pattern in HRNet~\cite{Wang-TPAMI-HRNet-2019}.
HRNet has shown a stronger capability 
among large models in position-sensitive problems, 
\eg semantic segmentation, human pose estimation, and object detection.
It remains unclear whether high resolution helps for small models.
We empirically show that the direct combination
outperforms ShuffleNet, MobileNet, and Small HRNet\footnote{Small HRNet is available at \url{https://github.com/HRNet/HRNet-Semantic-Segmentation}.
It simply reduces the depth and the width
of the original HRNet.}.

To further achieve higher efficiency, we introduce an efficient unit,
named \emph{conditional channel weighting},
performing information exchange across channels,
to replace the costly pointwise ($1\times 1$) convolution in a shuffle block.
The channel weighting scheme is very efficient:
the complexity is  
linear w.r.t the number of channels
and lower than
the quadratic time complexity for the pointwise convolution.
For example, with the multi-resolution features of $64\times64\times40$ and $32\times32\times80$,
the conditional channel weighting unit can reduce
the shuffle block's whole computation complexity
by $80\%$.

Unlike the regular convolutional kernel weights
learned as model parameters,
the proposed scheme weights
are conditioned on the input maps
and computed across channels
through a lightweight unit.
Thus, they contain the information 
in all the channel maps
and serve as a bridge
to exchange information  through channel weighting.
Furthermore,
we compute the weights
from parallel multi-resolution channel maps
that are readily available HRNet
so that the weights contain richer information
and are strengthened.
We call the resulting network, Lite-HRNet.

The experimental results show
that Lite-HRNet outperforms
the simple combination of shuffle blocks and HRNet
(which we call naive Lite-HRNet).
We believe that
the superiority is because the computational complexity reduction
is more significant
than the loss of information exchange
in the proposed conditional channel weighting scheme.

Our main contributions include:
\begin{compactitem}
	\vspace{.1cm}
	\item We simply apply the shuffle blocks to HRNet, leading a lightweight network naive Lite-HRNet.
	We empirically show superior performance
	over MobileNet, ShuffleNet, and Small HRNet.
	
	\vspace{.1cm}
	\item We present an improved efficient network, Lite-HRNet.
	The key point is that we introduce
	an efficient conditional channel weighting unit to replace the costly $1\times 1$ convolution in shuffle blocks,
	and the weights are computed across channels and resolutions.
	
	\vspace{.1cm}
	\item Lite-HRNet is the state-of-the-art in terms of complexity and accuracy trade-off on COCO and MPII human pose estimation and easily generalized to semantic segmentation task.
\end{compactitem}

\section{Related Work}
\label{sec:related-work}

\noindent\textbf{Efficient blocks for classification.}
Separable convolutions
and group convolutions
have been increasingly popular
in lightweight networks,
such as 
MobileNet~\cite{Howard-Arxiv-MobileNet-2017, Sandler-CVPR-MobileNetv2-2018, Howard-ICCV-Mobilenetv3-2019}, 
IGCV3~\cite{Sun-BMVC-IGCV3-2018},
and
ShuffleNet~\cite{Zhang-CVPR-Shufflenet-2018, Ma-ECCV-Shufflenetv2-2018}.
Xception~\cite{Chollet-CVPR-Xception-2017} 
and MobileNetV1~\cite{Howard-Arxiv-MobileNet-2017}
disentangle one normal convolution into
depthwise convolution and pointwise convolution.
MobileNetV2 
and IGCV3~\cite{Sun-BMVC-IGCV3-2018} 
further combine linear bottlenecks
that are about low-rank kernels.
MixNet~\cite{Tan-Arxiv-MixConv-2019} applies 
mixed kernels on the depthwise convolutions.
EfficientHRNet~\cite{Neff-ARXIV-EfficientHRNet-2020}
introduces the mobile convolutions into 
HigherHRNet~\cite{Cheng-CVPR-HigherHRNet-2020}.

The information across channels are blocked in group convolutions and depthwise convolutions. The pointwise convolutions are heavily used to address it but are very costly in lightweight network design.
To reduce the complexity, grouping $1\times 1$ convolutions
with channel shuffling~\cite{Zhang-CVPR-Shufflenet-2018, Ma-ECCV-Shufflenetv2-2018}
or interleaving~\cite{Zhang-ICCV-IGCV1-2017, Xie-CVPR-IGCV2-2018, Sun-BMVC-IGCV3-2018}
are used
to keep information exchange across channels.
Our proposed solution
is a lightweight manner performing information exchange
across channels
to replace costly $1\times 1$ convolutions.

\vspace{.1cm}
\noindent\textbf{Mediating spatial information loss.}
The computation complexity is positively related to spatial resolution.
Reducing the spatial resolution with mediating spatial information loss
is another way to improve efficiency.
Encoder-decoder architecture is used 
to recover the spatial resolution,
such as ENet~\cite{Paszke-Arxiv-ENet-2016} and SegNet~\cite{Badrinarayanan-PAMI-SegNet-2017}.
ICNet~\cite{Zhao-CVPR-PSPNet-2017} 
applies different computations to different resolution inputs
to reduce the whole complexity.
BiSeNet~\cite{Yu-ECCV-BiSeNet-2018, Yu-ARXIV-BiSeNetV2-2020}
decouples the detail information and context information 
with different lightweight sub-networks.
Our solution follows the high-resolution pattern in HRNet
to maintain the high-resolution representation 
through the whole process.

\vspace{.1cm}
\noindent\textbf{Convolutional weight generation
and mixing.}
Dynamic filter networks~\cite{Xu-NIPS-DFN-2016} dynamically 
generates the convolution filters 
conditioned on the input.
Meta-Network~\cite{Munkhdalai-MLS-MetaNet-2017} adopts a meta-learner to 
generate weights to 
learn cross-task knowledge.
CondINS~\cite{Tian-ECCV-CondINS-2020} and SOLOV2~\cite{Wang-NIPS-SOLOV2-2020}
apply this design to the instance segmentation task,
generating the parameters of the mask sub-network
for each instance.
CondConv~\cite{Yang-NIPS-CondConv-2019} and Dynamic Convolution~\cite{Chen-CVPR-DynamicConv-2020}
learn a series of weights 
to mix the corresponding convolution kernels 
for each sample, increasing the model capacity.

Attention mechanism~\cite{Hu-CVPR-SEnet-2017, Hu-NIPS-GENet-2018, Woo-ECCV-CBAM-2018, Yu-CVPR-DFN-2018} can be regarded as 
a kind of conditional weight generation.
SENet~\cite{Hu-CVPR-SEnet-2017} uses global information
to learn the weights 
to excite or suppress the channel maps.
GENet~\cite{Hu-NIPS-GENet-2018} expands on this by
gathering local information 
to exploit
the contextual dependencies.
CBAM~\cite{Woo-ECCV-CBAM-2018} exploits the channel and spatial attention
to refine the features.

The proposed conditional channel weighting scheme
can be, in some sense, regarded
as a conditional channel-wise $1\times 1$ convolution.
Besides its cheap computation,
we exploit an extra effect
and use the conditional weights
as the bridge to exchange information 
across channels.

\begin{figure}[t]
\footnotesize
\centering
(a)~\includegraphics[scale=.78]{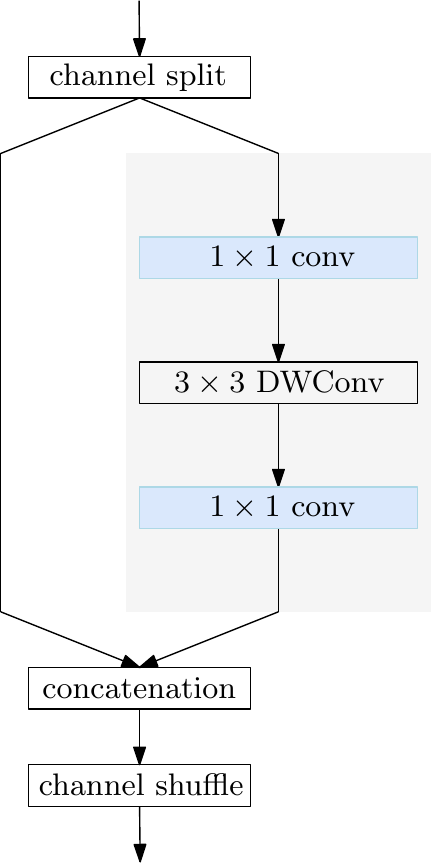}~~
(b)~\includegraphics[scale=.78]{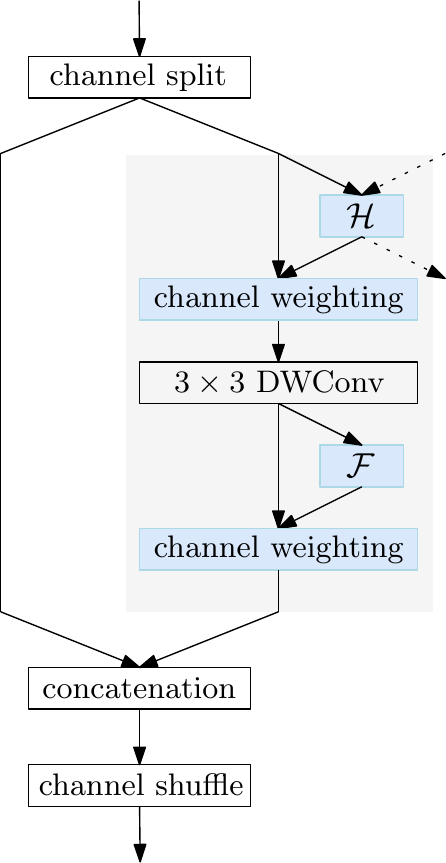}
\caption{\textbf{Building block.} (a) The shuffle block.
(b) Our conditional channel weighting block.
The dotted line indicates 
the representation from
other resolutions 
and the weights 
distributed to other resolutions. 
$\mathcal{H}$= cross-resolution weighting function. 
$\mathcal{F}$=spatial weighting function.}\
\label{fig:shufflenuintsandwccw}
\end{figure}

\vspace{.1cm}
\noindent\textbf{Conditional architecture.}
Different from normal networks,
conditional architecture can achieve 
dynamic width, depth, or kernels.
SkipNet~\cite{Wang-ECCV-Skipnet-2018} uses a gated network
to skip some convolutional
blocks to reduce complexity selectively.
Spatial Transform Networks~\cite{Jaderberg-NIPS-STN-2015} 
learn to warp the feature map conditioned on the input.
Deformable Convolution~\cite{Dai-ICCV-DCN-2017, Zhu-CVPR-DCVNV2-2019} learns
the offsets for the convolution kernels 
conditioned on each spatial location.

\section{Approach}
\label{sec:approach}

\subsection{Naive Lite-HRNet}

\noindent\textbf{Shuffle blocks.}
The shuffle block in ShuffleNet V2~\cite{Ma-ECCV-Shufflenetv2-2018}
first splits the channels into two partitions.
One partition passes through 
a sequence of 
$1\times 1$ convolution,
$3\times 3$ depthwise convolution,
and $1\times 1$ convolution,
and the output is concatenated
with the other partition.
Finally, the concatenated channels
are shuffled, 
as illustrated in Figure~\ref{fig:shufflenuintsandwccw} (a).

\begin{figure}[t]
\footnotesize
\centering
\includegraphics[width=1\linewidth]{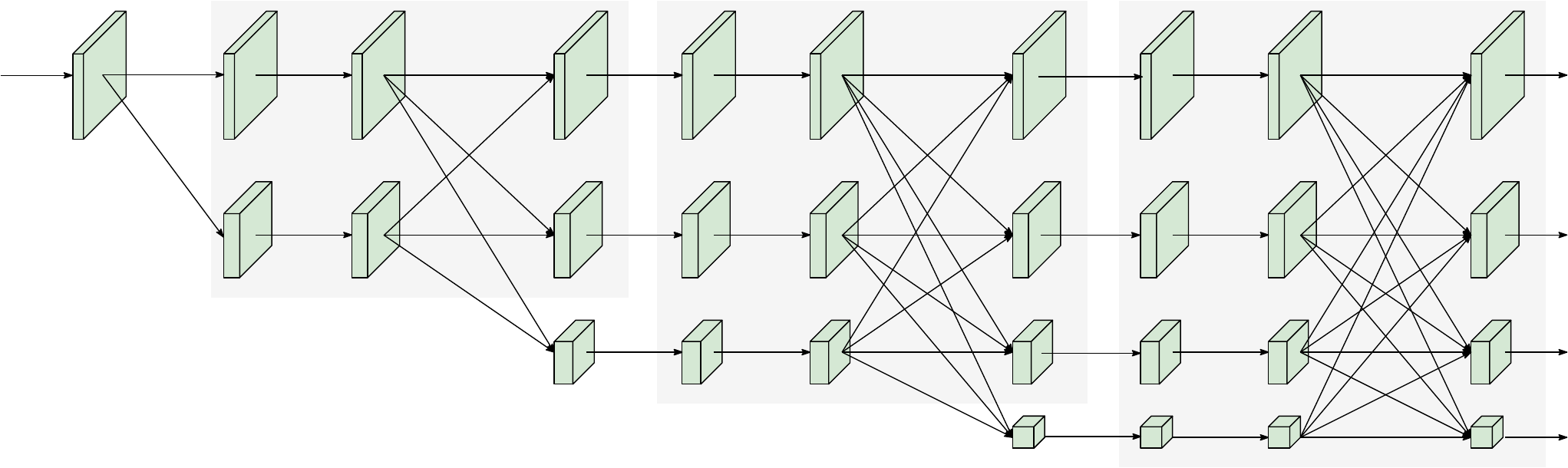}\\
\caption{\textbf{Illustration of the Small HRNet architecture.}
It consists of a high-resolution stem as the first stage, 
gradually adding high-to-low resolution streams as the main body. 
The main body has a sequence of stages, 
each containing parallel multi-resolution streams 
and repeated multi-resolution fusions. 
The details are given in Section~\ref{sec:approach}.}
\label{fig:hrnet}
\end{figure}

\vspace{.1cm}
\noindent\textbf{HRNet.}
The HRNet~\cite{Wang-TPAMI-HRNet-2019} starts 
from a high-resolution convolution stem as the first stage,
gradually adding high-to-low resolution streams one by one as new stages.
The multi-resolution streams are connected in parallel.
The main body consists of a sequence of stages.
In each stage, the information 
across resolutions 
is exchanged repeatedly.
We follow 
the Small HRNet design\footnote{\url{https://github.com/HRNet/HRNet-Semantic-Segmentation}}
and use fewer layers and smaller width
to form our network.
The stem of Small HRNet consists of two $3 \times 3$ convolutions with stride 2.
Each stage in the main body contains
a sequence of residual blocks and one multi-resolution fusion.
Figure~\ref{fig:hrnet} illustrates
the structure of Small HRNet.

\vspace{.1cm}
\noindent\textbf{Simple combination.}
We adopt the shuffle block to replace
the second $3 \times 3$ convolution in the stem of Small HRNet,
and replace all the normal residual blocks (formed with two $3 \times 3$ convolutions).
The normal convolutions
in the multi-resolution fusion
are replaced by the separable convolutions~\cite{Chollet-CVPR-Xception-2017},
resulting in a naive Lite-HRNet.

\subsection{Lite-HRNet}
\noindent\textbf{$1\times 1$ convolution is costly.}
The $1\times 1$ convolution 
performs a matrix-vector multiplication at each position:
\begin{align}
    \mathsf{Y} = \mathbf{W}\otimes\mathsf{X},
\end{align}
where $\mathsf{X}$ and $\mathsf{Y}$
are input and output maps,
and $\mathbf{W}$
is the $1\times 1$ convolutional kernel.
It serves a critical role of exchanging information across channels
as the shuffle operation and the depthwise convolution 
have no effect on information exchange across channels.

The $1\times 1$ convolution is of
quadratic time complexity ($\Theta(C^2)$) with respect to the number ($C$) of channels.
The $3\times 3$ depthwise convolution is of linear time complexity ($\Theta(9C)$\footnote{In terms of time complexity, the constant $9$ should be ignored.
We keep it for analysis convenience.}).
In the shuffle block,
the complexity of two $1\times 1$ convolutions
is much higher than
that of the depthwise convolution:
$\Theta(2C^2) > \Theta(9C)$, 
for the usual case $C>5$.
Table~\ref{tab:complexity_comparison} shows 
an example of the complexity comparison 
between $1\times1$ convolutions 
and depthwise convolutions.

\vspace{.1cm}
\noindent\textbf{Conditional channel weighting.}
We propose to use the element-wise weighting operation
to replace the $1\times 1$ convolution
in naive Lite-HRNet, which has $s$ branches in the $s$th stage.
The element-wise weighting operation
for the $s$th resolution branch is written as,
\begin{align}
    \mathsf{Y}_s = \mathsf{W}_s \odot \mathsf{X}_s,
\end{align}
where $\mathsf{W}_s$ is a weight map,
a $3$-d tensor of size $W_s\times H_s \times C_s$,
and $\odot$ is the element-wise multiplication operator. 

The complexity is linear with respect to
the channel number $\Theta(C)$,
and much lower than $1\times 1$ convolution
in the shuffle block. 

We compute the weights by using the channels for a single resolution 
and the channels across all the resolutions, as shown in Figure~\ref{fig:shufflenuintsandwccw}~(b),
and show that the weights play a role
of exchanging information across channels
and resolutions.

\vspace{.1cm}
\noindent\textbf{Cross-resolution weight computation.}
Considering the $s$-th stage,
there are $s$ parallel resolutions,
and $s$ weight maps $\mathsf{W}_1, \mathsf{W}_2,
\dots, \mathsf{W}_s$,
each for the corresponding resolution.
We compute the $s$ weight maps
from all the channels across resolutions
using a lightweight function $\mathcal{H}_s(\cdot)$,
\begin{align}
    (\mathsf{W}_1, \mathsf{W}_2, \dots, \mathsf{W}_s) = \mathcal{H}_s(\mathsf{X}_1, \mathsf{X}_2,\dots, \mathsf{X}_s), 
\end{align}
where $\{\mathsf{X}_1, \dots, \mathsf{X}_s\}$ are the input maps 
for the $s$ resolutions.
$\mathsf{X}_1$ corresponds to the highest resolution,
and $\mathsf{X}_s$
corresponds to the $s$-th highest resolution.

\begin{table*}[t]  
\footnotesize
\centering
\setlength{\tabcolsep}{10pt}
        \footnotesize
            \renewcommand{\arraystretch}{1.3}
\caption{\textbf{Structure of Lite-HRNet}. 
The stem contains one stride 2 $3\times3$ convolution and one shuffle block.
The main body has three stages, each of which has a sequence of modules.
Each module consists of two conditional channel weight blocks and one fusion block.
$N$ in Lite-HRNet-$N$ indicates the number of layers.
\emph{resolution branch} indicates this stage contains the feature stream of the corresponding resolution.
ccw = conditional channel weight. }
\label{tab:litehrnet}
\begin{tabular}{l|c|l|l|l|c|c|c}
\shline
\multirow{2}{*}{layer} & \multirow{2}{*}{output size} & \multicolumn{1}{c|}{\multirow{2}{*}{operator}}& \multirow{2}{*}{resolution branch} & \multicolumn{1}{c|}{\multirow{2}{*}{\#\emph{output\_channels}}} & \multirow{2}{*}{repeat} & \multicolumn{2}{c}{\#\emph{modules}} \\
\cline{7-8}
 & & &  & & & Lite-HRNet-18 & Lite-HRNet-30 \\
\shline
image & $256\times256$ & & $1\times$ & 3 & & & \\
\hline
\multirow{2}{*}{stem} & \multirow{2}{*}{$64\times64$} & conv2d & $2\times$ & 32 & 1 & \multirow{2}{*}{1} & \multirow{2}{*}{1} \\
\cline{3-6}
& & shuffle block & $4\times$ & 32 & 1 &  &  \\
\hline
\multirow{2}{*}{stage$_2$} & \multirow{2}{*}{$64\times64$} & ccw block & $4\times$ $8\times$ & 40, 80 & 2 & \multirow{2}{*}{2} & \multirow{2}{*}{3}  \\ 
\cline{3-6}
& & fusion block & $4\times$ $8\times$ & 40, 80 & 1 & & \\   
\hline                     
\multirow{2}{*}{stage$_3$} & \multirow{2}{*}{$64\times64$} & ccw block & $4\times$ $8\times$ $16\times$ & 40, 80, 160 & 2 & \multirow{2}{*}{4} & \multirow{2}{*}{8}  \\ 
\cline{3-6}
& & fusion block & $4\times$ $8\times$ $16\times$ & 40, 80, 160 & 1 & & \\   
\hline 
\multirow{2}{*}{stage$_4$} & \multirow{2}{*}{$64\times64$} & ccw block & $4\times$ $8\times$ $16\times$ $32\times$ & 40, 80, 160, 320 & 2 & \multirow{2}{*}{2} & \multirow{2}{*}{3}  \\ 
\cline{3-6}
& & fusion block & $4\times$ $8\times$ $16\times$ $32\times$ & 40, 80, 160, 320 & 1 & & \\   
\hline
$\operatorname{FLOPs}$ & & & &   & & 273.4M & 425.3M \\
\hline
$\operatorname{\#Params}$ & & & & & & 1.1M   & 1.8M \\
\shline
\end{tabular}
\end{table*}

\begin{table*}[t]  
\footnotesize
\centering
\setlength{\tabcolsep}{10pt}
        \footnotesize
            \renewcommand{\arraystretch}{1.3}
\caption{\textbf{Computational complexity comparison: $1\times1$ convolution \vs conditional channel weight}.
$\mathsf{X}_s \in \mathcal{R}^{H_s \times W_s \times C_s}$ are the input channel maps 
for the $s$ resolution, $\mathsf{X}_1$ corresponds to the highest resolution.
$N_s = H_s W_s$.
For example, the shape of $\mathsf{X}_1$ and $\mathsf{X}_2$ are $64 \times 64 \times 40$ and $32 \times 32 \times 80$, respectively.
single/cross-resolution=single/cross resolution information exchange.}
\label{tab:complexity_comparison}
\begin{tabular}{l|cc|c|c}
\shline
\multicolumn{1}{c|}{model} & single-resolution & cross-resolution &Theory Complexity & Example FLOPs \\
\shline
$1\times1$ convolution & \checkmark & &$\sum_1^s{N_s C_s^2}$ & 12.5M\\
$3\times3$ depthwise convolution & & & $\sum_1^s{9N_sC_s}$ & 2.1M \\
\hline
CCW w/ spatial weights & \checkmark & &$\sum_1^s{(2C_s^2+N_s C_s)}$ & 0.25M\\
CCW w/ multi-resolution weights &   & \checkmark &$2(\sum_1^s{C_s})^2 + \sum_1^s{N_s C_s}$ & 0.26M\\
\hline
CCW & \checkmark & \checkmark & $2(\sum_1^s{C_s})^2 + 2\sum_1^s{(C_s^2+N_s C_s)}$& 0.51M \\
\shline
\end{tabular}
\end{table*}

We implement the lightweight function $\mathcal{H}_s(\cdot)$
as following.
We perform adaptive average pooling ($\operatorname{AAP}$) 
on $\{\mathsf{X}_1, \mathsf{X}_2, \dots, \mathsf{X}_{s-1}\}$:
$\mathsf{X}_1' = \operatorname{AAP}(\mathsf{X}_1)$, 
$\mathsf{X}_2' = \operatorname{AAP}(\mathsf{X}_2)$, $\dots$, $\mathsf{X}_{s-1}' = \operatorname{AAP}(\mathsf{X}_{s-1})$,
in which the AAP pools any input size to a given output size $W_s \times H_s$.
Then we concatenate $\{\mathsf{X}_1', \mathsf{X}_2', \dots, \mathsf{X}_{s-1}'\}$
and $\mathsf{X}_s$ together,
followed by a sequence of $1 \times 1$ convolution, ReLU, 
$1 \times 1$ convolution, and sigmoid,
generating weight maps consisting of $s$ partitions,
$\mathsf{W}'_1, \mathsf{W}'_2, \dots, \mathsf{W}_s$
(each for one resolution):
\begin{align}
    (\mathsf{X}'_1, \mathsf{X}'_2,\dots, \mathsf{X}_s) \rightarrow &~\operatorname{Conv.}
\rightarrow \operatorname{ReLU}
\rightarrow \operatorname{Conv.} 
\rightarrow \operatorname{sigmoid} \nonumber \\
\rightarrow &~(\mathsf{W}'_1, \mathsf{W}'_2, \dots, \mathsf{W}'_s).
\label{eqn:crwc}
\end{align}
Here, the weights at each position for each resolution
depend on the channel feature at the same position
from the average-pooled multi-resolution channel maps.
This is why
we call the scheme as cross-resolution weight computation.
The $s-1$ weight maps, $\mathsf{W}_1', \mathsf{W}_2', \dots, \mathsf{W}'_{s-1}$, 
are upsampled to the corresponding resolutions,
outputting $\mathsf{W}_1, \mathsf{W}_2, \dots, \mathsf{W}_{s-1}$,
for the subsequent element-wise channel weighting.

We show that 
the weight maps serves
as a bridge 
for information exchange across channels
and resolutions. 
Each element of the weight vector $\mathbf{w}_{si}$ at the position $i$
(from the weight map $\mathsf{W}_s$)
receives the information from all the input channels 
of all the $s$ resolutions 
at the same pooling region,
which is easily verified from the operations in~Equation~\ref{eqn:crwc}.
Through such a weight vector,
each of the output channels at this position,
\begin{align}
    \mathbf{y}_{si} = \mathbf{w}_{si} \odot \mathbf{x}_{si},
\end{align}
receives the information from all the input channels
at the same position
across all the resolutions.
In other words,
the channel weighting scheme plays the role as well as the $1 \times 1$ convolution in terms of exchanging information.

On the other hand,
the function $\mathcal{H}_s(\cdot)$ is applied on the small resolution,
and thus the computation complexity is very light.
Table~\ref{tab:complexity_comparison} illustrates
that the whole unit has much lower complexity
than $1\times1$ convolution.

\vspace{.1cm}
\noindent\textbf{Spatial weight computation.}
For each resolution, we also compute the spatial weights which are homogeneous to spatial positions:
the weight vector $\mathbf{w}_{si}$
at all positions are the same.
The weights depend on all the pixels 
of the input channels in a single resolution:
\begin{align}
\mathbf{w}_{s} = \mathcal{F}_s(\mathsf{X}_s).
\end{align}
Here, the function $\mathcal{F}_s(\cdot)$ is implemented
as: $\mathsf{X}_s \rightarrow
\operatorname{GAP} \rightarrow \operatorname{FC}
\rightarrow \operatorname{ReLU}
\rightarrow \operatorname{FC}
\rightarrow \operatorname{sigmoid}
\rightarrow \mathbf{w}_{s}
$.
The global average pooling ($\operatorname{GAP}$) operator
serves as a role
of gathering the spatial information
from all the positions.

By weighting the channels with the spatial weights,
$\mathbf{y}_{si} = \mathbf{w}_{s} \odot \mathbf{x}_{si}$,
each element in the output channels
receives the contribution
from all the positions of all the input channels.
We compare the complexity between
$1\times1$ convolutions 
and 
conditional channel weighting unit in Table~\ref{tab:complexity_comparison}.

\begin{figure*}[t]
\footnotesize
\centering
\includegraphics[height=0.135\linewidth]{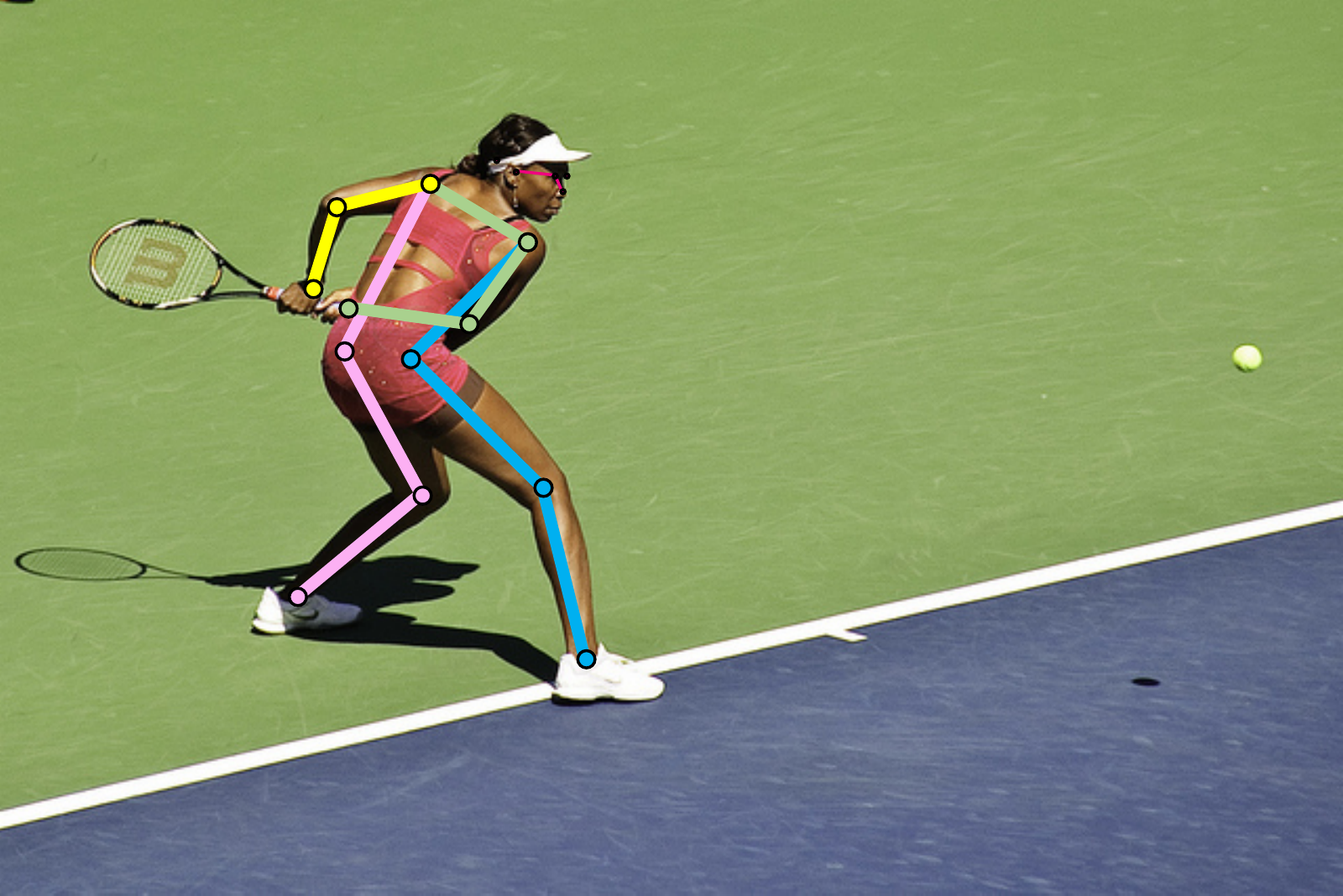}
\includegraphics[height=0.135\linewidth]{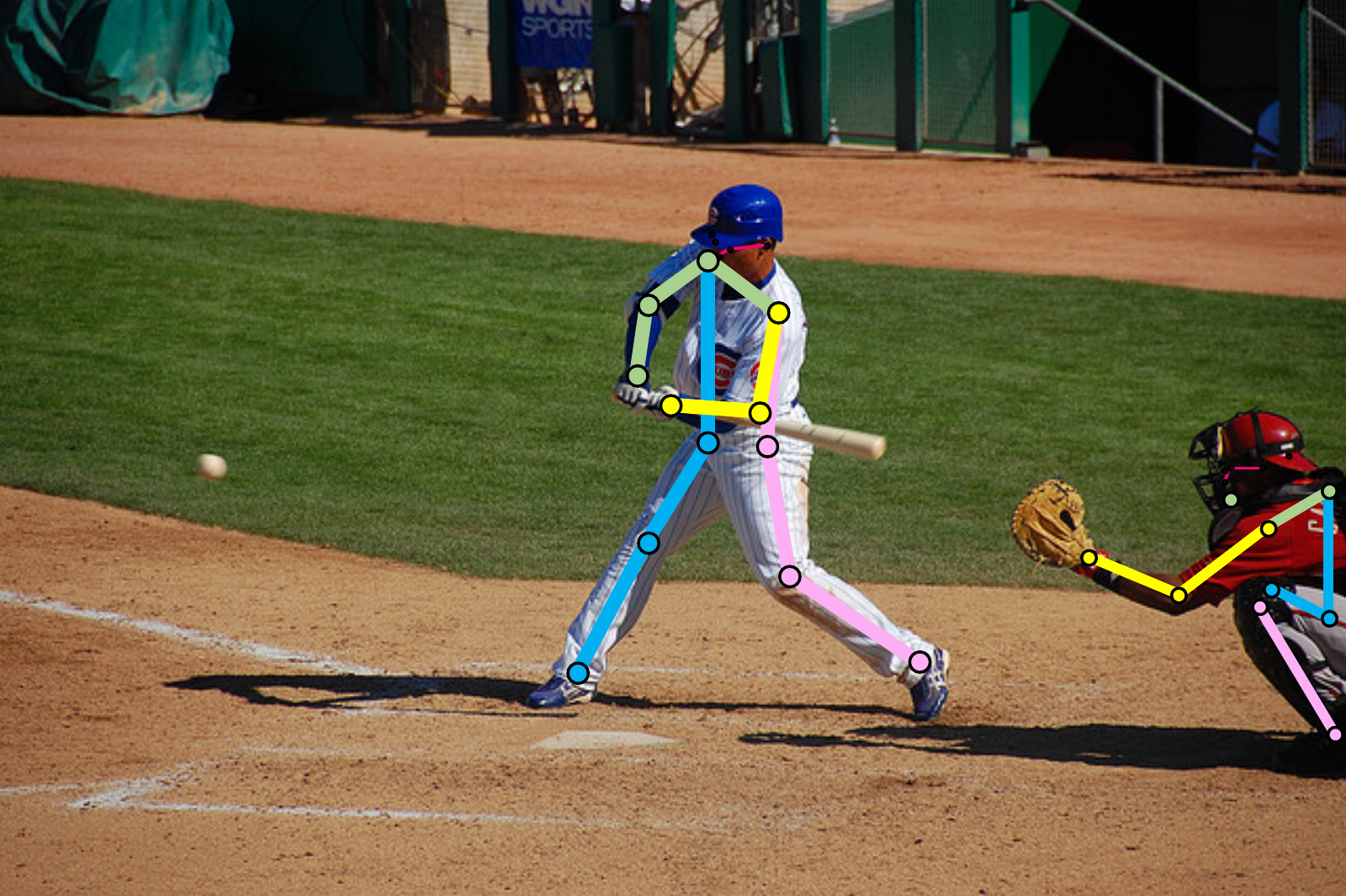}
\includegraphics[height=0.135\linewidth]{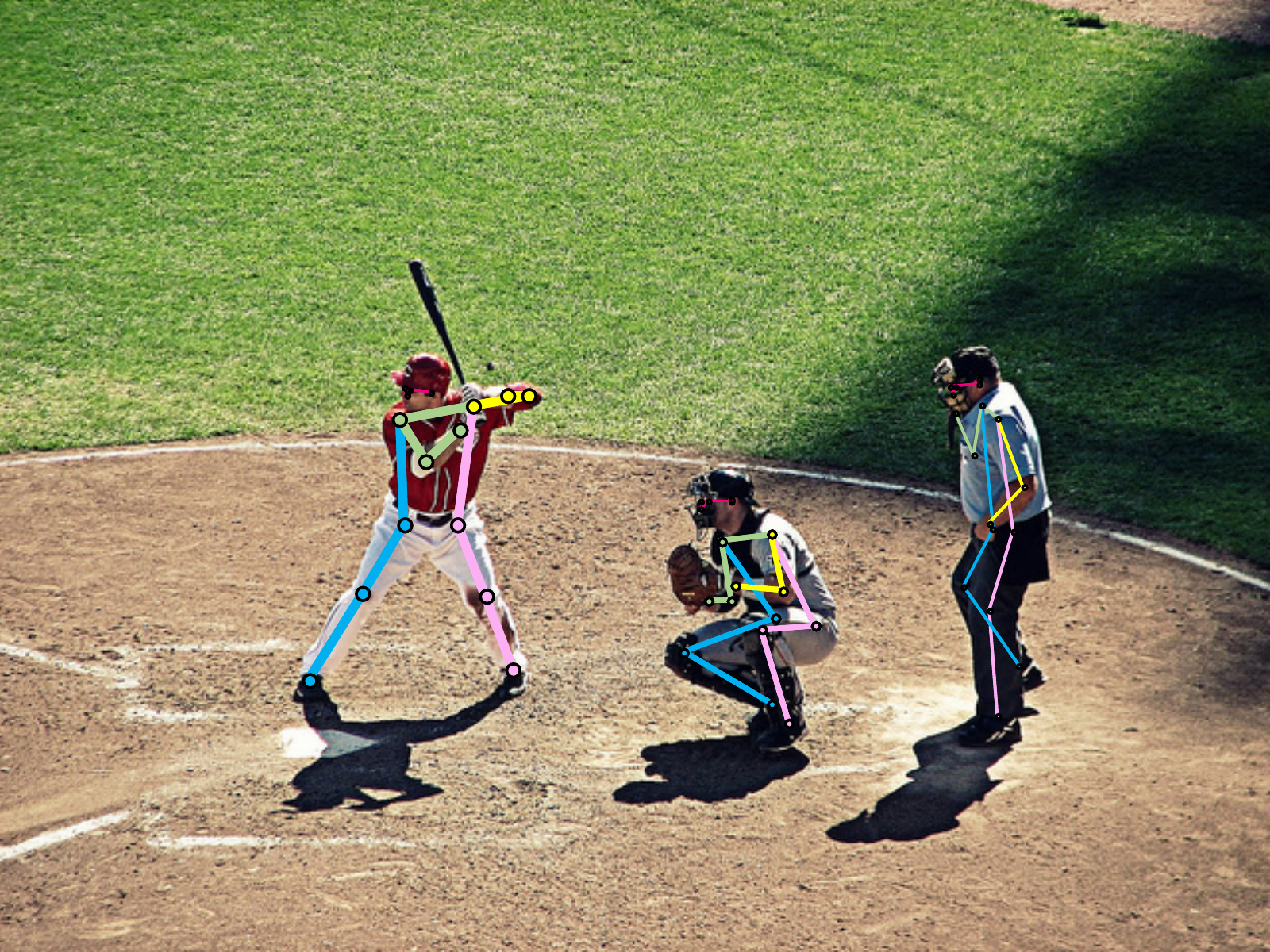}
\includegraphics[height=0.135\linewidth]{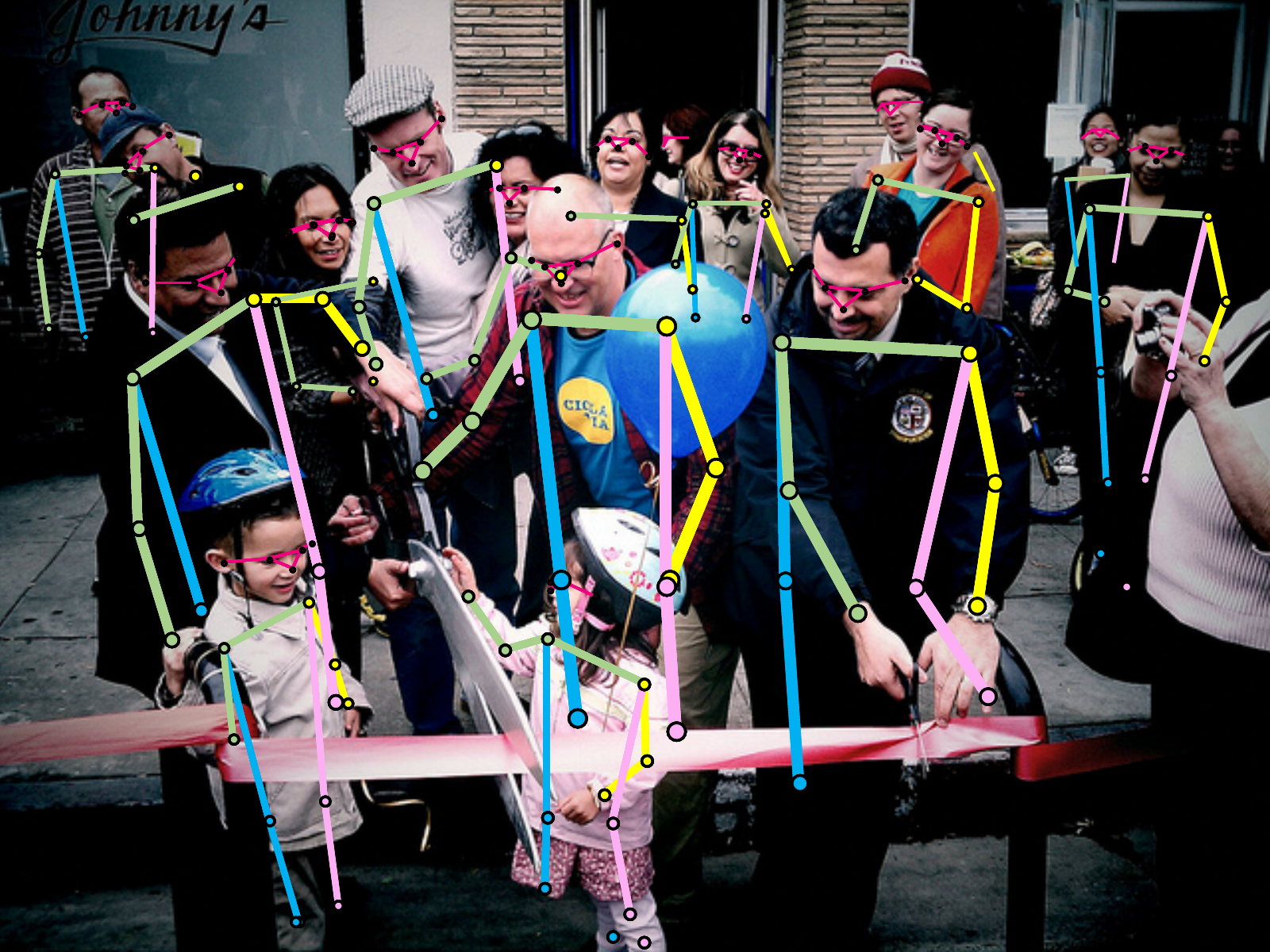} 
\includegraphics[height=0.135\linewidth]{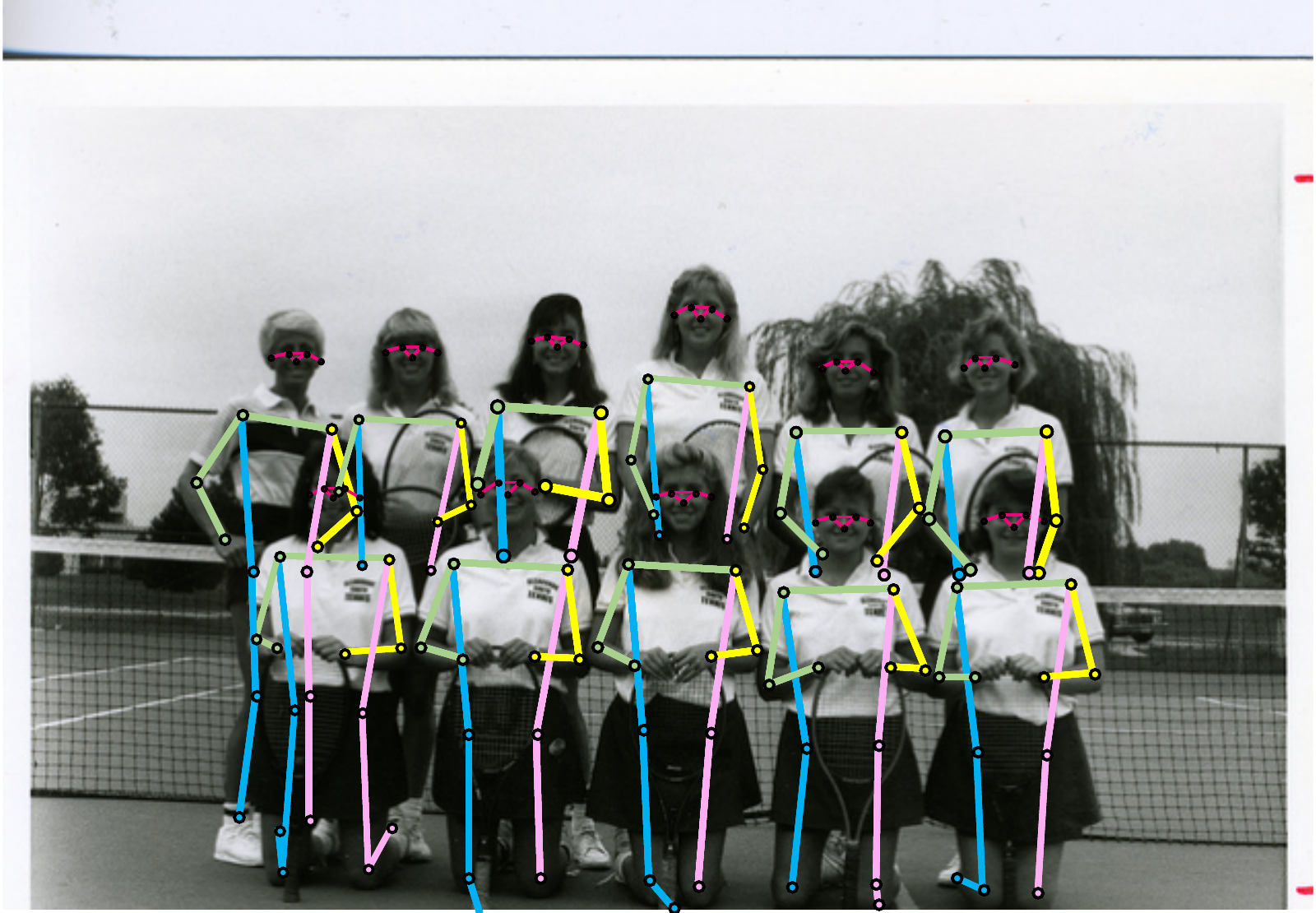}
\caption{\textbf{Example qualitative results on COCO pose estimation:}
containing viewpoint change, occlusion, and multiple persons.}
\label{fig:visualresults}
\end{figure*}

\begin{table*}[t]
\centering
\setlength{\tabcolsep}{5.0pt}
        \footnotesize
            \renewcommand{\arraystretch}{1.4}
            \caption{\textbf{Comparisons on the COCO \texttt{val} set.}
            pretrain = pretrain the backbone on ImageNet. 
\#Params and FLOPs are calculated 
for the pose estimation network, and 
those for human detection and keypoint grouping are not included.}
\label{tab:coco_val}
\begin{tabular}{l|l|c|r|c|c|cccccc}
\shline
\multicolumn{1}{c|}{model} & \multicolumn{1}{c|}{backbone} &  pretrain & input size & $\operatorname{\#Params}$ & $\operatorname{GFLOPs}$ & 
$\operatorname{AP}$ & $\operatorname{AP}^{50}$ & $\operatorname{AP}^{75}$ & $\operatorname{AP}^{M}$ & $\operatorname{AP}^{L}$ & $\operatorname{AR}$ \\
\shline
\multicolumn{11}{l}{\emph{Large networks}}\\
\hline
$8$-stage Hourglass~\cite{Newell-ECCV-Hourglass-2016} & $8$-stage Hourglass & N &  $256 \times 192$ & $25.1$M & $14.3$ &
$66.9$&$-$&$-$&$-$&$-$&$-$\\ 
CPN~\cite{Chen-CVPR-CPN-2018}& ResNet-50~\cite{He-CVPR-ResNet-2016} & Y & $256 \times 192$ & $27.0$M & $6.20$ &
$68.6$&$-$&$-$&$-$&$-$&$-$\\ 
SimpleBaseline~\cite{Xiao-ECCV-SimpleBaseline-2018} & ResNet-50 & Y & $256\times192$  &$34.0$M &$8.90 $
&${70.4}$ & ${88.6}$&${78.3}$&${67.1}$&${77.2}$&${76.3}$\\
HRNetV$1$~\cite{Wang-TPAMI-HRNet-2019} & HRNetV$1$-W$32$ & N & $256\times 192$&  $28.5$M & $7.10$ &
$73.4$&$89.5$&$80.7$&$70.2$&$80.1$&$78.9$  \\
DARK~\cite{Zhang-CVPR-DARK-2020} & HRNetV1-W48 & Y & $128\times96$ & $63.6$M & $3.6$ & 71.9 & 89.1 & 79.6 & 69.2 & 78.0 & 77.9 \\
\hline
\multicolumn{11}{l}{\emph{Small networks}}\\
\hline
MobileNetV2 1$\times$~\cite{Sandler-CVPR-MobileNetv2-2018} & MobileNetV2 & Y & $256\times192$ & $9.6$M & $1.48$ & 64.6 & 87.4 & 72.3 & 61.1 & 71.2 & 70.7 \\
MobileNetV2 1$\times$& MobileNetV2 & Y & $384\times288$ & $9.6$M & $3.33$ & 67.3 & 87.9 & 74.3 & 62.8 & 74.7 & 72.9 \\
ShuffleNetV2 1$\times$~\cite{Ma-ECCV-Shufflenetv2-2018} & ShuffleNetV2 & Y & $256\times192$ & $7.6$M & $1.28$ & 59.9 & 85.4 & 66.3 & 56.6 & 66.2 & 66.4 \\
ShuffleNetV2 1$\times$& ShuffleNetV2 & Y & $384\times288$ & $7.6$M & $2.87$ & 63.6 & 86.5 & 70.5 & 59.5 & 70.7 & 69.7 \\
Small HRNet             & HRNet-W16 & N & $256\times192$ & $1.3$M  & $0.54$ & 55.2 & 83.7 & 62.4 & 52.3 & 61.0 & 62.1 \\
Small HRNet             & HRNet-W16 & N & $384\times288$ & $1.3$M  & $1.21$ & 56.0 & 83.8 & 63.0 & 52.4 & 62.6 &  62.6 \\
DY-MobileNetV2 1$\times$~\cite{Chen-CVPR-DynamicConv-2020}& DY-MobileNetV2 & Y & $256\times192$ & $16.1$M  & $1.01$ & 68.2 & 88.4 & 76.0 & 65.0 & 74.7 &  74.2 \\
DY-ReLU 1$\times$~\cite{Chen-ECCV-DyReLU-2020}& MobileNetV2 & Y & $256\times192$ & $9.0$M  & $1.03$ & 68.1 & 88.5 & 76.2 & 64.8 & 74.3 & $-$ \\
\hline
Lite-HRNet & Lite-HRNet-18 & N & $256\times192$ & $1.1$M & $0.20$ & 64.8 & 86.7 & 73.0 & 62.1 & 70.5 & 71.2 \\
Lite-HRNet & Lite-HRNet-18 & N & $384\times288$ & $1.1$M & $0.45$ & 67.6  & 87.8 & 75.0 & 64.5 & 73.7 & 73.7  \\
Lite-HRNet & Lite-HRNet-30 & N & $256\times192$ & $1.8$M & $0.31$ & 67.2 & 88.0 & 75.0 & 64.3 & 73.1 & 73.3 \\
Lite-HRNet & Lite-HRNet-30 & N & $384\times288$ & $1.8$M & $0.70$ & 70.4 & 88.7 & 77.7 & 67.5 & 76.3 & 76.2  \\
\shline
\end{tabular}
\end{table*}

\vspace{.1cm}
\noindent\textbf{Instantiation.}
The Lite-HRNet consists of 
a high-resolution stem and
the main body 
to maintain the high-resolution representation.
The stem has one $3\times 3$ convolution with stride 2 
and a shuffle block, as the first stage.
The main body has a sequence of modularized modules.
Each module consists of two 
conditional channel weighting blocks
and 
one multi-resolution fusion.
Each resolution branch's 
channel dimensions are $C$, $2C$, $4C$, and $8C$, respectively.
Table~\ref{tab:litehrnet} describes
the detailed structures.

\vspace{.1cm}
\noindent\textbf{Connection.}
The conditional channel weighting scheme 
shares the same philosophy
to the conditional convolutions~\cite{Yang-NIPS-CondConv-2019}, dynamic filters~\cite{Xu-NIPS-DFN-2016}, and squeeze-excite-network~\cite{Hu-CVPR-SEnet-2017}.
Those works learn the convolution kernels
or the mixture weights 
by sub-network conditioned on the input features
for increasing the model capacity.
Our method instead exploits an extra effect
and uses the weights learned from all the channels
as a bridge
to exchange information across channels and resolutions. 
It can replace costly $1\times 1$ convolutions
in lightweight networks.
Besides, we introduce multi-resolution information
to boost weight learning.

\section{Experiments}
We evaluate our approach on 
two human pose estimation datasets, 
COCO~\cite{Lin-COCO-2014} and 
MPII~\cite{Andriluka-CVPR-MPII-2014}.
Following the state-of-the-art top-down framework, 
our approach estimates $K$ heatmaps 
to indicate the keypoint location confidence.
We perform a comprehensive ablation 
on COCO and report the comparisons 
with other methods on both datasets.

\subsection{Setting}

\begin{table*}[t]
\centering
\setlength{\tabcolsep}{6pt}
        \footnotesize
            \renewcommand{\arraystretch}{1.3}
            \caption{\textbf{Comparisons on the COCO \texttt{test-dev} set.}
            \#Params and FLOPs are calculated 
            for the pose estimation network, and 
            those for human detection and keypoint grouping are not included.
            }
            \label{tab:coco_test_dev}
\begin{tabular}{l|l|c|c|c|lllllc}
\shline
\multicolumn{1}{c|}{model} & \multicolumn{1}{c|}{backbone} & input size & $\operatorname{\#Params}$ & $\operatorname{GFLOPs}$ &
$\operatorname{AP}$ & $\operatorname{AP}^{50}$ & $\operatorname{AP}^{75}$ & $\operatorname{AP}^{M}$ & $\operatorname{AP}^{L}$ & $\operatorname{AR}$\\
\shline
\multicolumn{11}{l}{\emph{Large networks}}\\
\hline
Mask-RCNN~\cite{He-ICCV-MaskRCNN-2017} & ResNet-50-FPN & $-$ &$-$& $-$
& $63.1$ & $87.3$&$68.7$&$57.8$&$71.4$&$-$\\
G-RMI~\cite{Papandreou-CVPR-GRMI-2017} & ResNet-101 & $353\times257$ &$42.6$M& $57.0$
&$64.9$ & $85.5$&$71.3$&$62.3$&$70.0$&$69.7$\\
Integral Pose Regression~\cite{Sun-ECCV-Integral-2018} & ResNet-101 & $256\times256$ &$45.0$M& $11.0$
&$67.8$ & $88.2$&$74.8$&$63.9$&$74.0$&$-$\\
CPN~\cite{Chen-CVPR-CPN-2018} & ResNet-Inception& $384\times288$ &$-$& $-$
& $72.1$ & $91.4$&$80.0$&$68.7$&$77.2$&$78.5$\\
RMPE~\cite{Fang-ICCV-Rmpe-2017} & PyraNet~\cite{Yang-ICCV-PyraNet-2017} & $320\times256$ &$28.1$M& $26.7$
&$72.3$ & $89.2$&$79.1$&$68.0$&$78.6$&$-$\\
SimpleBaseline~\cite{Xiao-ECCV-SimpleBaseline-2018} & ResNet-152&$384\times288$  &$68.6$M& $35.6$
&${73.7}$ & ${91.9}$&${81.1}$&${70.3}$&${80.0}$&${79.0}$\\

HRNetV$1$~\cite{Wang-TPAMI-HRNet-2019} & HRNetV$1$-W$32$& $384\times 288$ &$28.5$M&$16.0$ &$74.9$&$92.5$&$82.8$&$71.3$&$80.9$&$80.1$\\
HRNetV$1$~\cite{Wang-TPAMI-HRNet-2019} & HRNetV$1$-W$48$& $384\times 288$ &$63.6$M&$32.9$ &$75.5$&$92.5$&$83.3$&$71.9$&$81.5$&$80.5$\\
DARK~\cite{Zhang-CVPR-DARK-2020} & HRNetV$1$-W$48$ & $384\times288$ & $63.6$M & $32.9$ & 76.2 & 92.5 & 83.6 & 72.5 & 82.4 & 81.1\\
\hline

\multicolumn{11}{l}{\emph{Small networks}}\\
\hline
MobileNetV2 1$\times$& MobileNetV2 & $384\times288$ & $9.8$M & $3.33$ & 66.8 & 90.0 & 74.0 & 62.6 & 73.3 & 72.3 \\
ShuffleNetV2 1$\times$& ShuffleNetV2 & $384\times288$ & $7.6$M & $2.87$ & 62.9 & 88.5 & 69.4 & 58.9 & 69.3 & 68.9 \\
Small HRNet & HRNet-W16 & $384\times288$ & $1.3$M & $1.21$ & 55.2 & 85.8 & 61.4 & 51.7 & 61.2 & 61.5 \\
\hline
Lite-HRNet & Lite-HRNet-18   & $384\times288$ & $1.1$M & $0.45$ & 66.9 & 89.4 & 74.4 & 64.0 & 72.2 & 72.6  \\
Lite-HRNet & Lite-HRNet-30   & $384\times288$ & $1.8$M & $0.70$ & 69.7 & 90.7 & 77.5 & 66.9 & 75.0 & 75.4  \\
\shline
\end{tabular}
\end{table*}

\noindent\textbf{Datasets.}
COCO~\cite{Lin-COCO-2014} has over $200K$ images 
and $250K$ person instances with 17 keypoints.
Our models are trained on \texttt{train2017} dataset 
(includes $57K$ images and $150K$ person instances) 
and validated on \texttt{val2017} (includes $5K$ images) 
and \texttt{test-dev2017} (includes $20K$ images). 

The MPII Human Pose dataset~\cite{Andriluka-CVPR-MPII-2014} 
contains around $25K$ images with full-body pose annotations 
taken from real-world activities.
There are over $40K$ person instances, 
split $12K$ instances for testing, 
and others for training.

\begin{table}[t]  
\footnotesize
\centering
\setlength{\tabcolsep}{10pt}
        \footnotesize
            \renewcommand{\arraystretch}{1.3}
\caption{\textbf{Comparisons on the MPII \texttt{val} set.}
The FLOPs is computed with 
the input size $256\times256$.}
\label{tab:mpii_comparison}
\begin{tabular}{l|c|c|c}
\shline
\multicolumn{1}{c|}{model} & $\operatorname{\#Params}$ & $\operatorname{GFLOPs}$ & $\operatorname{PCKh}$ \\
\shline
MobileNetV2 1$\times$   & $9.6$M & $1.97$ & 85.4 \\
MobileNetV3 1$\times$& $8.7$M & $1.82$ & 84.3  \\
ShuffleNetV2 1$\times$  & $7.6$M & $1.70$  & 82.8  \\
Small HRNet-W$16$ 	  & $1.3$M & $0.72$ & 80.2 \\
\hline
Lite-HRNet-$18$  &  $1.1$M & $0.27$   & 86.1  \\
Lite-HRNet-$30$ &  $1.8$M & $0.42$   & 87.0  \\
\shline
\end{tabular}
\end{table}

\begin{figure*}[t]
\centering
{\footnotesize (a)}~\begin{tikzpicture}[baseline]
\pgfplotsset{set layers, compat=1.3, 
every axis/.append style={
font=\footnotesize,
}
}
\begin{axis}[
	footnotesize,
	scale only axis,
	ybar,
	enlargelimits=0.15,
	symbolic x coords={
	MBV2, SFV2, SHR, WLH-18, LH-18, LH-30},
	x post scale=1.4,
	xtick=data,
    nodes near coords, 
	nodes near coords align={vertical},
    x tick label style={rotate=45,anchor=east},
    axis y line*=left,
    every axis y label/.style={
at={(ticklabel cs:1.07)},rotate=0,anchor=near ticklabel, xshift=3em
},
	ylabel=\color{mayablue}{GFLOPs},
]
\addplot[fill=mayablue, draw=mayablue]
	coordinates{
	(MBV2, 1.48)
	(SFV2, 1.28)
	(SHR, 0.54)
	(WLH-18, 0.3)
	(LH-18, 0.20)
	(LH-30, 0.31)
	};

\end{axis}
\begin{axis}[
	footnotesize,
	scale only axis,
	enlargelimits=0.15,
	symbolic x coords={
	MBV2, SFV2, SHR, WLH-18, LH-18, LH-30},
	x post scale=1.4,
	xtick=data,
    nodes near coords, 
	nodes near coords align={vertical},
	axis x line=none,
	axis y line*=right,
	every axis y label/.style={
at={(ticklabel cs:1.07)},rotate=0,anchor=near ticklabel, xshift=-3em
},
	ymin=55, ymax=70,
	ylabel=\color{mangotango}{mAP},
	grid=major,
]
	\addplot[mangotango, mark=pentagon*, mark size=3pt, line width=1pt, smooth]
	coordinates{
	(MBV2, 64.6)
	(SFV2, 59.9)
	(SHR, 55.2)
	(WLH-18, 65.7)
	(LH-18, 64.8)
	(LH-30, 67.2)
	};
\end{axis}
\end{tikzpicture}%
~~
{\footnotesize (b)}~\begin{tikzpicture}[baseline]
\pgfplotsset{set layers, compat=1.3, 
every axis/.append style={
font=\footnotesize,
}
}
\begin{axis}[
    footnotesize,
	scale only axis,
	ybar,
	enlargelimits=0.15,
	symbolic x coords={
	MBV2, SFV2, SHR, WLH-18, LH-18, LH-30},
	x post scale=1.4,
	xtick=data,
    nodes near coords, 
	nodes near coords align={vertical},
    x tick label style={rotate=45,anchor=east},
    axis y line*=left,
    every axis y label/.style={
at={(ticklabel cs:1.07)},rotate=0,anchor=near ticklabel, xshift=3em
},
	ylabel=\color{mayablue}{GFLOPs},
]

\addplot[fill=mayablue, draw=mayablue]
	coordinates{
	(MBV2, 1.97)
	(SFV2, 1.7)
	(SHR, 0.72)
	(WLH-18, 0.41)
	(LH-18, 0.27)
	(LH-30, 0.42)
	};

\end{axis}
\begin{axis}[
	footnotesize,
	scale only axis,
	enlargelimits=0.15,
	symbolic x coords={
	MBV2, SFV2, SHR, WLH-18, LH-18, LH-30},
	x post scale=1.4,
	xtick=data,
    nodes near coords, 
	nodes near coords align={vertical},
	axis x line=none,
	axis y line*=right,
	every axis y label/.style={
at={(ticklabel cs:1.07)},rotate=0,anchor=near ticklabel, xshift=-3em
},
	ymin=78, ymax=90,
	ylabel=\color{mangotango}{PCKh},
	grid=major,
]
	\addplot[mangotango, mark=pentagon*, mark size=3pt, line width=1pt, smooth]
	coordinates{
	(MBV2, 85.4)
	(SFV2, 82.8)
	(SHR, 80.2)
	(WLH-18, 86.8)
	(LH-18, 86.1)
	(LH-30, 87.0)
	};
\end{axis}
\end{tikzpicture} \\
\caption{\textbf{Illustration of the complexity and accuracy comparison on the COCO \texttt{val} and MPII \texttt{val} sets.} 
(a) Comparison on COCO \texttt{val} with $256\times 192$ input size.
(b) Comparison on MPII \texttt{val} with $256\times 256$ input size.
MBV2= MobileNet V2.
SFV2= ShuffleNet V2.
SHR= Small HRNet-W16.
(W)LH= (Wider Naive) Lite-HRNet.
}
\label{fig:naivelitehrnetvssmallhernet}
\end{figure*}
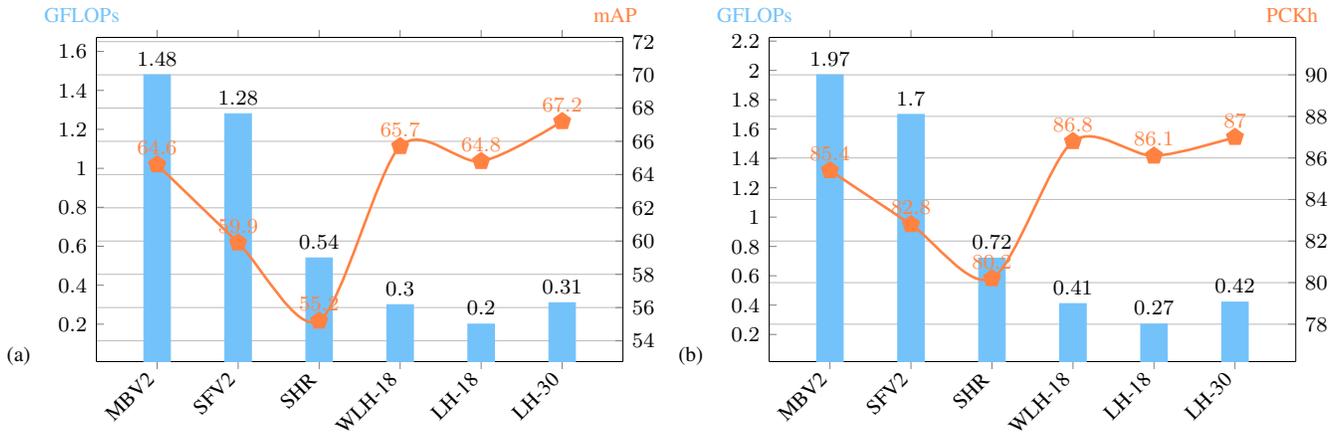

\begin{table*}[t]
\centering
\setlength{\tabcolsep}{7.5pt}
        \footnotesize
            \renewcommand{\arraystretch}{1.3}
\caption{\textbf{Ablation about conditional channel weight \vs
$1\times 1$ convolutions} on the COCO \texttt{val} and MPII \texttt{val} sets.
The input size of COCO is $256\times 192$, 
while $256\times 256$ for MPII.
Wider NLite-NRNet = wider naive Lite-HRNet.}
\begin{tabular}{l|c|c|cccc|c|c}
\shline
\multicolumn{1}{c|}{\multirow{2}{*}{model}} & \multirow{2}{*}{$\operatorname{\#Params}$} & \multicolumn{5}{c|}{COCO} & \multicolumn{2}{c}{MPII} \\
\cline{3-9}
 & & $\operatorname{MFLOPs}$ & $\operatorname{AP}$ & $\operatorname{AP}^{50}$ & $\operatorname{AP}^{75}$ & $\operatorname{AR}$
 & $\operatorname{MFLOPs}$ & $\operatorname{PCKh}$\\
\shline
Small HRNet-W16             & $1.3$M  & $551.7$ & 55.2 & 83.7 & 62.4 & 62.1 & $735.5$ & 80.2 \\
Naive Lite-HRNet-$18$ & $0.7$M & $194.8$ & 62.5 & 85.4 & 69.6 & 68.8 & $259.6$ & 85.3\\
Wider Naive Lite-HRNet-$18$ & $1.3$M & $311.1$ & 65.7 & 87 & 73.3 & 71.8  & $418.7$ & 86.8\\
\hline
Wider NLite-HRNet-$18$ (\emph{one $1\times 1$ conv. dropped}) & $1.1$M & $248.4$ & 63.6 & 86.1 & 70.7 & 69.8 & $331.0$ & 86.0\\
Wider NLite-HRNet-$18$ (\emph{two $1\times 1$ conv. dropped}) & $0.9$M & $188.9$ & 61.3 & 85.3 & 68.7 & 67.7 & $251.7$ & 85.3\\
\hline
Lite-HRNet-$18$ & $1.1$M & $205.2$ & 64.8 & 86.7 & 73.0 & 71.2 & $273.4$ & 86.1 \\
Lite-HRNet-$30$ & $1.8$M & $319.2$ & 67.2 & 88.0 & 75.0 & 73.3 & $425.3$ & 87.0 \\
\shline
\end{tabular}
\label{tab:ablation_CCWvs11}
\end{table*}

\vspace{.1cm} 
\noindent\textbf{Training.}
The network is trained on 8 NVIDIA V100 GPUs
with mini-batch size 32 per GPU.
We adopt Adam optimizer with an initial learning rate of $2e^{-3}$.

The human detection boxes are expanded to have a fixed aspect ratio of 4: 3,
and then crop the box from the images.
The image size is resized to $256\times192$ or 
$384\times288$ for COCO, 
and $256\times256$ for MPII.
Each image will go through 
a series of data augmentation operations, 
containing random rotation ($[\ang{-30}, \ang{30}]$),
random scale ($[0.75, 1.25]$), and random flipping 
for both datasets and 
additional half body data augmentation
for COCO.

\vspace{.1cm} 
\noindent\textbf{Testing.}
For COCO, following~\cite{Xiao-ECCV-SimpleBaseline-2018, Chen-CVPR-CPN-2018, Papandreou-CVPR-GRMI-2017}, we adopt the two-stage top-down paradigm 
(detect the person instance via a person detector 
and predict keypoints) with the person detectors 
provided by SimpleBaseline~\cite{Xiao-ECCV-SimpleBaseline-2018}. 
For MPII, we adopt the 
standard testing strategy 
to use the provided person boxes.
We estimate the heat maps via a post-gaussian filter
and average the original and flipped images' predicted heat maps. 
A quarter offset in the direction from the highest response to the second-highest response is applied to obtain each keypoint location.

\vspace{.1cm} 
\noindent\textbf{Evaluation.}
We adopt the OKS-based mAP metric on COCO, 
where $\operatorname{OKS}$ (Object Keypoint Similarity) defines 
the similarity between different human poses.
We report standard average precision and recall scores:
$\operatorname{AP}$ (the mean of $\operatorname{AP}$ scores at 10 positions, 
$\operatorname{OKS}=0.50,0.55,\dots,0.90,0.95$), 
$\operatorname{AP}^{50}$ ($\operatorname{AP}$ at $\operatorname{OKS}=0.50$), 
$\operatorname{AP}^{75}$, $\operatorname{AR}$ and $\operatorname{AR}^{50}$.
For MPII, we use the standard metric $\operatorname{PCKH}@0.5$ (head-normalized probability of correct keypoint) to evaluate the performance.

\subsection{Results}

\noindent\textbf{COCO \texttt{val}.}
The results of our method
and other state-of-the-art methods
are reported in Table~\ref{tab:coco_val}.
Our Lite-HRNet-30, trained from scratch with 
the $256\times192$ input size, 
achieves $67.2$ AP score,
outperforming other light-weight methods.
Compared to MobileNetV2,
Lite-HRNet improves AP
by $2.6$ points with only
$20\%$ GFLOPs
and parameters.
Compared to ShuffleNetV2,
our Lite-HRNet-18 and Lite-HRNet-30
achieve $4.9$ and $7.3$ points gain,
respectively.
The complexity of our network
is much lower than ShuffleNetV2.
Compared to Small HRNet-W16,
Lite-HRNet improves over $10$ AP points.
Compared to large networks, 
\eg Hourglass and CPN,
our networks achieve comparable
AP score with far lower complexity.

With the input size $384\times 288$, 
our Lite-HRNet-18 and Lite-HRNet-30
achieve $67.6$ and $70.4$ AP, 
respectively.
Due to the efficient
conditional channel weighting,
Lite-HRNet achieves a better balance
between accuracy and computational
complexity, as shown in 
Figure~\ref{fig:naivelitehrnetvssmallhernet}~(a).
Figure~\ref{fig:visualresults}
shows the visual results on COCO
from Lite-HRNet-30.

\vspace{.1cm}
\noindent\textbf{COCO \texttt{test-dev}.}
Table~\ref{tab:coco_test_dev}
reports the comparison results 
of our networks 
and
other state-of-the-art methods.
Our Lite-HRNet-30 achieves
$69.7$ AP score.
It is significantly better than
the small networks, 
and is more efficient
in terms of GFLOPs and parameters.
Compared to the large networks,
our Lite-HRNet-30 outperforms
Mask-RCNN~\cite{He-ICCV-MaskRCNN-2017}, 
G-RMI~\cite{Papandreou-CVPR-GRMI-2017}, 
and Integral Pose Regression~\cite{Sun-ECCV-Integral-2018}.
Although there is a performance gap with 
some large networks,
our networks have far lower 
GFLOPs and parameters.

\begin{table*}[t]
\centering
\setlength{\tabcolsep}{7pt}
        \footnotesize
            \renewcommand{\arraystretch}{1.3}

\caption{\textbf{Ablation about spatial and multi-resolution weights.} on the COCO \texttt{val} and MPII \texttt{val} sets.
The input size of COCO is $256\times 192$, 
while $256\times 256$ for MPII.
CCW=conditional channel weight computation,
Wider NLite-NRNet = wider naive Lite-HRNet.}
\label{tab:ablation_ccw}
\begin{tabular}{l|c|c|cccc|c|c}
\shline
\multicolumn{1}{c|}{\multirow{2}{*}{model}} & \multirow{2}{*}{$\operatorname{\#Params}$} & 
\multicolumn{5}{c|}{COCO} & \multicolumn{2}{c}{MPII} \\
\cline{3-9}
& & $\operatorname{MFLOPs}$ & $\operatorname{AP}$ & $\operatorname{AP}^{50}$ & $\operatorname{AP}^{75}$ & $\operatorname{AR}$ 
& $\operatorname{MFLOPs}$ & 
$\operatorname{PCKh}$ \\
\shline
Wider NLite-HRNet-$18$ (\emph{two $1\times 1$ conv. dropped}) & $0.9$M & $188.9$ & 61.3 & 85.3 & 68.7 & 67.7 & $251.7$ & 85.3\\
\hline
Lite-HRNet-$18$ (\emph{CCW only w/ spatial weights}) & $0.9$M & $190.6$ & 62.6 & 85.8 & 69.8 & 69.1 & $254.0$ & 85.4 \\
Lite-HRNet-$18$ (\emph{CCW only w/ multi-resolution weights}) & $0.9$M & $203.5$ & 63.0 & 85.7 & 70.5 & 69.4 & $271.1$ & 85.8 \\
\hline
Lite-HRNet-$18$ & $1.1$M & $205.2$ & 64.8 & 86.7 & 73.0 & 71.2 & $273.4$ & 86.1 \\
Lite-HRNet-$30$ & $1.8$M & $319.2$ & 67.2 & 88.0 & 75.0 & 73.3 & $425.3$ & 87.0 \\
\shline
\end{tabular}
\end{table*}

\vspace{.1cm}
\noindent\textbf{MPII \texttt{val}.}
Table~\ref{tab:mpii_comparison}
reports the results of our networks 
and other lightweight networks.
Our Lite-HRNet-18 achieves 
better accuracy with much lower
GFLOPs than 
MobileNetV2,
MobileNetV3,
ShuffleNetV2,
Small HRNet-W16.
With increasing the model size, 
as Lite-HRNet-30,
the improvement gap becomes larger.
Our Lite-HRNet-30 achieves
$87.0$ $\operatorname{PCKh}@0.5$,
improving 
MobileNetV2, MobileNetV3, ShuffleNetV2
and Small HRNet-W16 by 
$1.6$, $2.7$, $4.2$, and 
$6.8$ points, respectively.
Figure~\ref{fig:naivelitehrnetvssmallhernet}~(b)
shows the comparison of accuracy and complexity.

\subsection{Ablation Study}
We perform ablations on two datasets: COCO and MPII, 
and report the results on the validation sets.
The input size is $256\times192$ for COCO,
and $256 \times 256$ for MPII.

\vspace{.1cm}
\noindent\textbf{Naive Lite-HRNet \vs Small HRNet.}
We empirically study that
the shuffle blocks combined into HRNet
improve performance.
Figure~\ref{fig:naivelitehrnetvssmallhernet}
shows the comparison to Small HRNet-W$16$\footnote{Available from \url{https://github.com/HRNet/HRNet-Semantic-Segmentation})}.
We can see that naive Lite-HRNet achieves higher AP scores
with lower computation complexity.
On COCO \texttt{val},
naive Lite-HRNet improves AP over the Small HRNet-W16 
by 7.3 points, 
and the GFLOPs and parameters are less than half.
When increasing to similar parameters as wider naive Lite-HRNet,
the improvement is enlarged to 10.5 points,
as shown in Figure~\ref{fig:naivelitehrnetvssmallhernet} (a).
On MPII \texttt{val},
naive Lite-HRNet outperforms 
the Small HRNet-W16 by 5.1 points,
while the wider network outperforms 6.6 points,
as illustrated in Figure~\ref{fig:naivelitehrnetvssmallhernet} (b).

\vspace{.1cm}
\noindent\textbf{Conditional channel weighting \vs 
$1\times 1$ convolution.}
We compare the performance
between $1\times 1$ convolution (wider naive Lite-HRNet)
and conditional channel weighting (Lite-HRNet).
We simply remove one or two 
$1\times 1$ convolutions
in the shuffle blocks in wider naive Lite-HRNet.

Table~\ref{tab:ablation_CCWvs11}
shows the studies on the COCO \texttt{val}
and MPII \texttt{val} sets.
$1\times1$ convolutions can exchange 
the information across channels,
important to representation learning.
On COCO \texttt{val},
dropping two $1\times1$ convolutions
leads to $4.4$ AP points decrease 
for 
wider naive Lite-HRNet, 
and also reduces almost $40\%$ FLOPs.

Our conditional channel weighting
improves by $3.5$ AP points 
over dropping two $1\times1$ convolutions 
with only increasing $16$M FLOPs.
The AP score is comparable with 
the wider naive Lite-HRNet
by using only $65\%$ FLOPs.
Increasing the depth of Lite-HRNet 
leads to $1.5$ AP improvements
with similar FLOPs as
wider naive Lite-HRNet
and slightly larger \#parameters than wider naive Lite-HRNet.
The observations on MPII \texttt{val}
are consistent
(see Table~\ref{tab:ablation_CCWvs11}).
The AP improvement is because 
that our lightweight weighting operations 
can make the network capacity improved, 
by exploring the multi-resolution information 
using cross-resolution channel weighting
and deepening the network,
if taking similar FLOPs with naive version.

\vspace{.1cm}
\noindent\textbf{Spatial and multi-resolution weights.}
We empirically study how spatial weights
and multi-resolution weights 
influence the performance,
as shown in Table~\ref{tab:ablation_ccw}.

On COCO \texttt{val},
the spatial weights achieve
1.3 AP increase, 
and the multi-resolution weights
obtain 1.7 point gain.
The FLOPs of both operations are cheap.
With both spatial and cross-resolution weights,
our network improves by $3.5$ points. 
Table~\ref{tab:ablation_ccw}
reports the consistent improvements 
on MPII \texttt{val}.
These studies validate 
the efficiency and effectiveness of 
the spatial and cross-resolution weights.

We conduct the experiments by changing the arrangement order 
of the spatial weighting 
and cross-resolution weighting,
which achieves similar performance.
The experiments 
with only two spatial weights
or two cross-resolution weights,
lead to 
an almost $0.3$ drop.

\subsection{Application to Semantic Segmentation}

\noindent\textbf{Dataset.}
Cityscapes~\cite{Cityscapes} 
includes 30 classes 
and 19 of them are used 
for semantic segmentation task. 
The dataset contains 2,975, 500, and 1,525 
finely-annotated images for training, 
validation, and test sets, respectively. 
In our experiments, 
we only use the fine annotated images.

\vspace{.1cm}
\noindent\textbf{Training.} 
Our models are trained from scratch 
with the SGD algorithm~\cite{Krizhevsky-NIPS-Imagenet}.
The initial rate is set to $1e^{-2}$ 
with a ``poly'' learning rate strategy~\cite{Yu-CVPR-CPNet-2020, Yu-ECCV-RepGraph-2020} 
with a multiplier 
of $(1-\frac{iter}{max\_iters})^{0.9}$ each iteration. 
The total iterations are 160K 
with 16 batch size, and the weight decay is $5e^{-4}$.
We randomly horizontally flip, scale ($[0.5, 2]$), 
and crop the input images 
to a fixed size ($512\times 1024$) for training. 

\vspace{.2cm}
\noindent\textbf{Results.}
We do not adopt testing tricks, 
\eg sliding-window and multi-scale evaluation, 
beneficial to performance improvement but time-consuming.
Table~\ref{tab:seg:cityscapes}
shows that
Lite-HRNet-18 achieves $72.8\%$ mIoU
with only $1.95$ GFLOPs
and
Lite-HRNet-30 achieves $75.3\%$ mIoU
with $3.02$ GFLOPs,
outperforming the hand-crafted methods~\cite{Paszke-Arxiv-ENet-2016, Zhao-ECCV-ICNet-2018, Yu-ECCV-BiSeNet-2018, Li-CVPR-DFANet-2019, Zhuang-CVPR-ShelfNet-2018, Sandler-CVPR-MobileNetv2-2018, Wang-TPAMI-HRNet-2019, Zhou-ECCV-MobileNext-2020}
and NAS-based methods~\cite{Zhang-CVPR-CAS-2019, Lin-CVPR-GAS-2020, Howard-ICCV-Mobilenetv3-2019, Li-CVPR-POP-2019}
, and comparable with SwiftNetRN-18~\cite{Orsic-CVPR-SwiftNet-2019}  that is far computationally intensive (104 GFLOPs).

\begin{table}[t]
\centering
\setlength{\tabcolsep}{1.5pt}
\footnotesize
\renewcommand{\arraystretch}{1.3}
\caption{\textbf{Segmentation results on
	Cityscapes.}
	P = pretrain the backbone on ImageNet. 
	$^*$ indicates the complexity is estimated 
	from the original paper.}
\label{tab:seg:cityscapes}
\begin{tabular}{l|l|cc|c|cc}
\shline
\multicolumn{1}{c|}{model} & \multicolumn{1}{c|}{P} & $\operatorname{\#Params}$  & $\operatorname{GFLOPs}$ & resolution & \textit{val} & \textit{test} \\ \shline
  \multicolumn{7}{l}{\emph{Hand-crafted networks}}\\
  \hline
ICNet \cite{Zhao-ECCV-ICNet-2018} & Y & $-$ & $28.3$  & $1024\times 2048$ & $67.7$ & $69.5$ \\
BiSeNetV1 A \cite{Yu-ECCV-BiSeNet-2018}& Y & $5.8$M & $14.8$ & $768\times 1536$ & $69.0$ & $68.4$ \\
BiSeNetV1 B \cite{Yu-ECCV-BiSeNet-2018}& Y & $49.0$M & $55.3$ & $768\times 1536$ & $74.8$ & $74.7$ \\
DFANet A' \cite{Li-CVPR-DFANet-2019}& Y & $7.8$M & $1.7$ & $512\times 1024$ & $-$ & $70.3$ \\
SwiftNet \cite{Orsic-CVPR-SwiftNet-2019}& Y & $11.8$M & $26.0$  & $512\times 1024$ & $70.2$ & $-$ \\
SwiftNet \cite{Orsic-CVPR-SwiftNet-2019}& Y & $11.8$M & $104$  & $1024\times 2048$ & $75.4$ & $75.5$ \\
Fast-SCNN \cite{Poudel-Arxiv-FastSCNN-2019}& N & $-$ & $-$  & $1024\times 2048$ & $68.6$ & $68.0$ \\
ShelfNet \cite{Zhuang-CVPR-ShelfNet-2018}& Y & $-$ & $36.9$  & $1024\times 2048$ & $-$ & $74.8$ \\
BiSeNetV2 Small~\cite{Yu-ARXIV-BiSeNetV2-2020} & N &  $-$ & 21.15 & $512 \times 1024$ & 73.4 & 72.6 \\
MoibleNeXt~\cite{Zhou-ECCV-MobileNext-2020}& Y & $4.5$M & $10.1^{*}$ & $1024 \times 2048$ & $75.5$ & $-$ \\
MobileNet V2 0.5~\cite{Sandler-CVPR-MobileNetv2-2018}  & Y & $0.3$M & $3.73$ & $512\times1024$ & 68.6 & $-$ \\
HRNet-W16~\cite{Wang-TPAMI-HRNet-2019} 	                 & Y & $2.0$M & $7.8$ & $512\times1024$ & 68.6 & $-$\\
  \hline
  \multicolumn{7}{l}{\emph{NAS-based networks}}\\
  \hline
CAS \cite{Zhang-CVPR-CAS-2019}& Y & $-$ & $-$ & $768\times 1536$ & $71.6$ & $70.5$ \\
DF1-Seg-d8 \cite{Li-CVPR-POP-2019} & Y & $-$ & $-$ & $1024\times 2048$ & $72.4$ & $71.4$ \\
FasterSeg \cite{Chen-ICLR-FasterSeg-2020}& Y & $4.4$M & $28.2$  & $1024\times 2048$ & $73.1$ & $71.5$ \\
GAS \cite{Lin-CVPR-GAS-2020} & Y & $-$ & $-$ & $769\times1537$ & $-$ & 71.8 \\
MobileNetV3 \cite{Howard-ICCV-Mobilenetv3-2019}& Y & $1.5$M & $9.1$ & $1024\times 2048$ & $72.4$ & $72.6$ \\
MobileNet V3-Small& Y & $0.5$M & $2.7$ & $512\times1024$ & 68.4 & 69.4\\
\hline
Lite-HRNet-$18$      & N & $1.1$M & $1.95$ & 512$\times$1024 & 73.8 & 72.8\\
Lite-HRNet-$30$    & N & $1.8$M & $3.02$ & 512$\times$1024 & 76.0 & 75.3 \\
\shline	
\end{tabular}
\end{table}

\section*{Acknowledgements}
This work is supported by the National Natural Science Foundation of China (No. 61433007 and 61876210).

\clearpage

{\small
\bibliographystyle{ieee_fullname}
\bibliography{reference}
}

\end{document}